\documentclass[10pt,conference]{IEEEtran}
\IEEEoverridecommandlockouts
\usepackage{cite}
\usepackage{amsmath,amssymb,amsfonts}

\usepackage{algorithm}
\usepackage{algpseudocode}

\usepackage{graphicx}
\usepackage{textcomp}
\usepackage{xcolor}
\usepackage{epsfig}
\usepackage{tcolorbox}
\usepackage{multirow}
\usepackage[singlelinecheck=false]{caption}
\usepackage{setspace}

\usepackage{bbm}
\usepackage{color}

\newtheorem{definition}{Definition}

\def\BibTeX{{\rm B\kern-.05em{\sc i\kern-.025em b}\kern-.08em
    T\kern-.1667em\lower.7ex\hbox{E}\kern-.125emX}}
\begin{document}

\bibliographystyle{plain}

\title{HSCoNAS: Hardware-Software Co-Design of Efficient DNNs via Neural Architecture Search

\thanks{This work is partially supported by the Ministry of Education, Singapore, under its Academic Research Fund Tier 2 (MOE2019-T2-1-071) and Tier 1 (MOE2019-T1-001-072), and partially supported by Nanyang Technological University, Singapore, under its NAP (M4082282) and SUG (M4082087).}
}
\author{
\vspace{0.13cm}
Xiangzhong Luo$^1$, Di Liu$^2$, Shuo Huai$^3$, Weichen Liu$^4$  \\
\textit{$^{1, 3, 4}$School of Computer Science and Engineering, Nanyang Technological University, Singapore}  \\
\textit{$^{2, 3, 4}$HP-NTU Digital Manufacturing Corporate Lab, Nanyang Technological University, Singapore}  \\
\textit{(xiangzho001$^1$, shuo001$^3$)@e.ntu.edu.sg, (liu.di$^2$, liu$^4$)@ntu.edu.sg}
}

\makeatletter
\patchcmd{\@maketitle}
  {\addvspace{0.5\baselineskip}\egroup}
  {\addvspace{-1\baselineskip}\egroup}
  {}
  {}
\makeatother

\linespread{0.956}

\maketitle
\vspace{-2cm}

\begin{abstract}
In this paper, we present a novel multi-objective hardware-aware neural architecture search (NAS) framework, namely HSCoNAS, to automate the design of deep neural networks (DNNs) with high accuracy but low latency upon target hardware. To accomplish this goal, we first propose an effective hardware performance modeling method to approximate the runtime latency of DNNs on target hardware, which will be integrated into HSCoNAS to avoid the tedious on-device measurements. Besides, we propose two novel techniques, \textit{i.e.}, dynamic channel scaling to maximize the accuracy under the specified latency and progressive space shrinking to refine the search space towards target hardware as well as alleviate the search overheads. These two techniques jointly work to allow HSCoNAS to perform fine-grained and efficient explorations. Finally, an evolutionary algorithm (EA) is incorporated to conduct the architecture search. Extensive experiments on ImageNet are conducted upon diverse target hardware, \textit{i.e.}, GPU, CPU, and edge device to demonstrate the superiority of HSCoNAS over recent state-of-the-art approaches. 
\end{abstract}


\section{Introduction}
\label{sec:introduction}

Deep neural networks (DNNs) have become the \textit{de facto} engine of artificial intelligence (AI). Over the past few years, DNNs have achieved remarkable success in a wide range of real-world applications, such as person re-identification \cite{luo2020reid}, autonomous driving \cite{hao2019nais}, intelligent IoT \cite{hao2019fpga}, \textit{etc}. However, to pursue competitive accuracy, DNNs are evolving deeply with more layers as well as widely with more channels, thereby incurring the \textit{computational gap} between complicated DNNs and resource-limited hardware like edge devices, which are deemed as the key computing platform for future AI \cite{liu2020survey, luo2020edgenas}. Nonetheless, designing resource-efficient DNNs for less capable hardware still remains challenging since hardware-aware DNNs need to be small and fast, yet still accurate. 

\begin{figure}[t]
    \begin{center}
        \includegraphics[width=0.80\columnwidth]{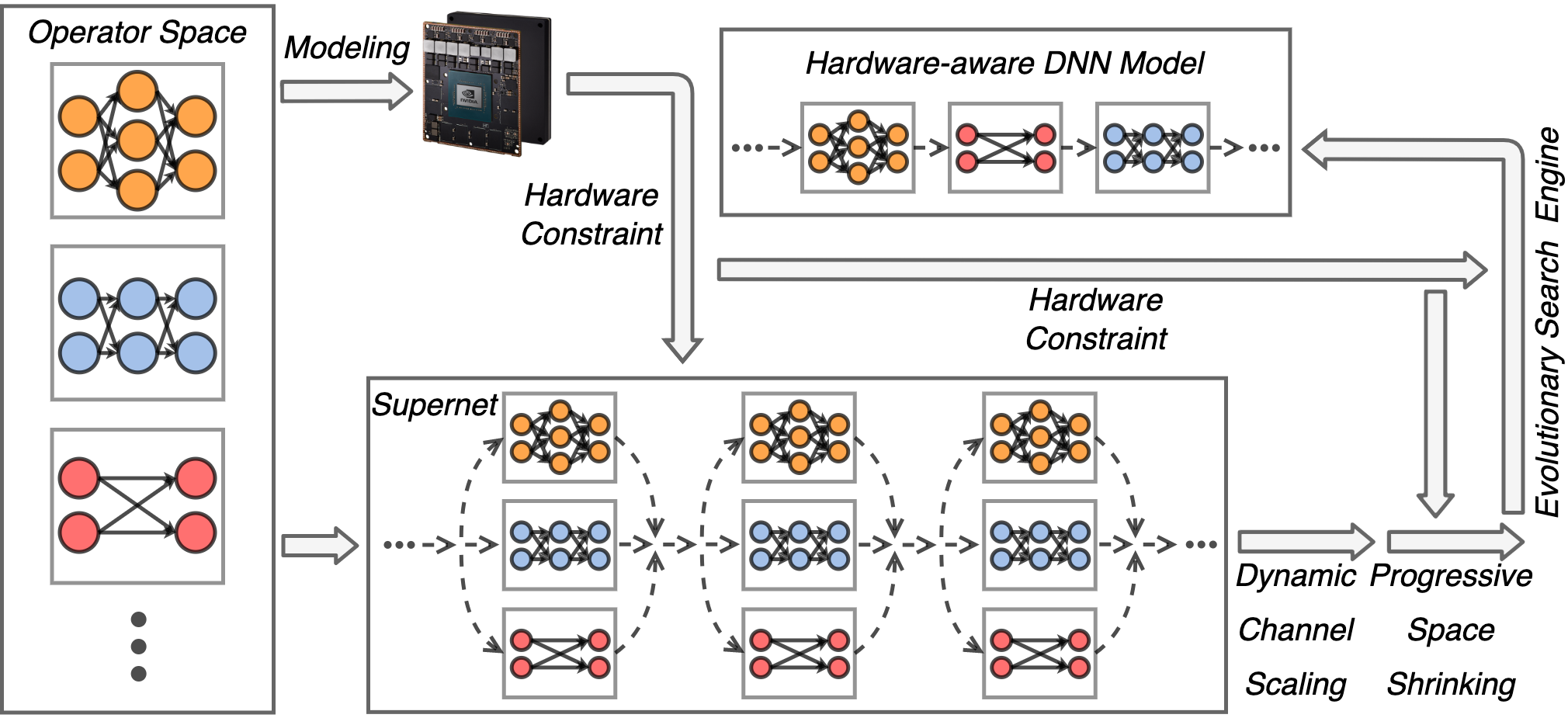}
        \vspace{-0.35cm}
    \end{center}
    \caption{Overview of the proposed HSCoNAS framework.}
\vspace{-0.5cm}
\label{fig:overview}
\end{figure}

To tackle the above \textit{computational gap}, unlike previous hardware-agnostic methods, we propose an efficient and unified \underline{h}ardware-\underline{s}oftware \underline{co}-design \underline{NAS} framework, namely HSCoNAS, to automatically design efficient DNNs with high accuracy but low latency upon diverse target hardware. The overview of HSCoNAS is illustrated in Fig. \ref{fig:overview}. Our main contributions are as follows:
\begin{itemize}
\vspace{-0.1cm}
\setlength{\itemsep}{0pt}
\setlength{\parskip}{0pt}
\setlength{\parsep}{0pt}
    \item [1)] 
    We propose a hardware performance modeling method to approximate the runtime latency of DNNs upon target hardware while introducing negligible overheads.
    \item [2)] 
    We formulate a multi-objective NAS approach, \textit{i.e.}, HSCoNAS. Besides, we propose a novel dynamic channel scaling scheme to enable the channel-level explorations. Also, we present a progressive space shrinking to refine the search space towards target hardware, followed by an evolutionary algorithm to perform efficient search. 
    \item [3)] 
    We perform extensive experiments on ImageNet with three hardware devices, \textit{i.e.}, GPU, CPU, and edge device, which demonstrate the superiority of HSCoNAS.
\vspace{-0.1cm}
\end{itemize}

\section{Preliminaries and Problem Formulation}
\label{sec:preliminaries}

\subsection{Preliminaries}
\label{sec:nas}

NAS has been regarded as a promising alternative to automate the design of competitive DNNs. Prior NAS works \cite{pham2018efficient, tan2019mnasnet, liu2018darts} usually construct an over-parameterized network with $L$ layers, namely supernet $\mathcal{N}$, to ease the search of optimal neural architectures. As illustrated in Fig. \ref{fig:overview}, the supernet can be formulated as a directed acyclic graph (DAG) based on a set of $K$ operators, \textit{i.e.}, $\mathcal{O}=\{op_i\}_{i=1}^K$, where each layer has $K$ different operators. Note the operator can be the basic convolution, pooling, or building blocks from manually-designed DNNs like ShuffleNetV2 \cite{ma2018shufflenet} and MobileNetV2 \cite{sandler2018mobilenetv2}. Finally, an architecture candidate $arch$ can be sampled from the supernet by selecting one operator for each layer, \textit{i.e.}, $arch=\{op^l\}_{l=1}^L \in \mathcal{N}=\{\mathcal{O}^l\}_{l=1}^L$. Note we set $L=20$ and $K=5$ across this work. Thus, once the supernet is well trained, we can evaluate architecture candidates (\textit{i.e.}, subgraphs) with inherited weights from the supernet by means of the weight-sharing technique \cite{pham2018efficient}, thereby avoiding the considerable overheads of training vast stand-alone DNNs.

\subsection{Problem Formulation}
\label{sec:problem-formulation}

In deep learning driven platforms, they may trade latency for higher accuracy, and vice versa. Thus, to accomplish flexibility, we present a multi-objective formula to achieve the trade-off between accuracy and latency as follows:
\begin{equation}
    {\small \mathop{\mathrm{maximize}}\limits_{arch \in \mathcal{A}} \,\, ACC(arch) + \beta \times \Big |\frac{LAT(arch)}{T} - 1 \Big|}
    \label{eq:relaxed-objective}
\end{equation}
where $\mathcal{A}$ denotes the search space. $ACC(\cdot)$ and $LAT(\cdot)$ denote accuracy on target task and runtime latency on target hardware, respectively. $T$ is the specified latency constraint on target hardware. $\beta < 0$ is the trade-off coefficient. For simplicity, we denote the above objective as $\mathcal{F}(arch, T)$. Therefore, we can either penalize the architecture with high latency or with low accuracy, thereby achieving an effective trade-off between accuracy and latency. 

\begin{figure}[t]
    \begin{center}
        \includegraphics[width=0.80\columnwidth]{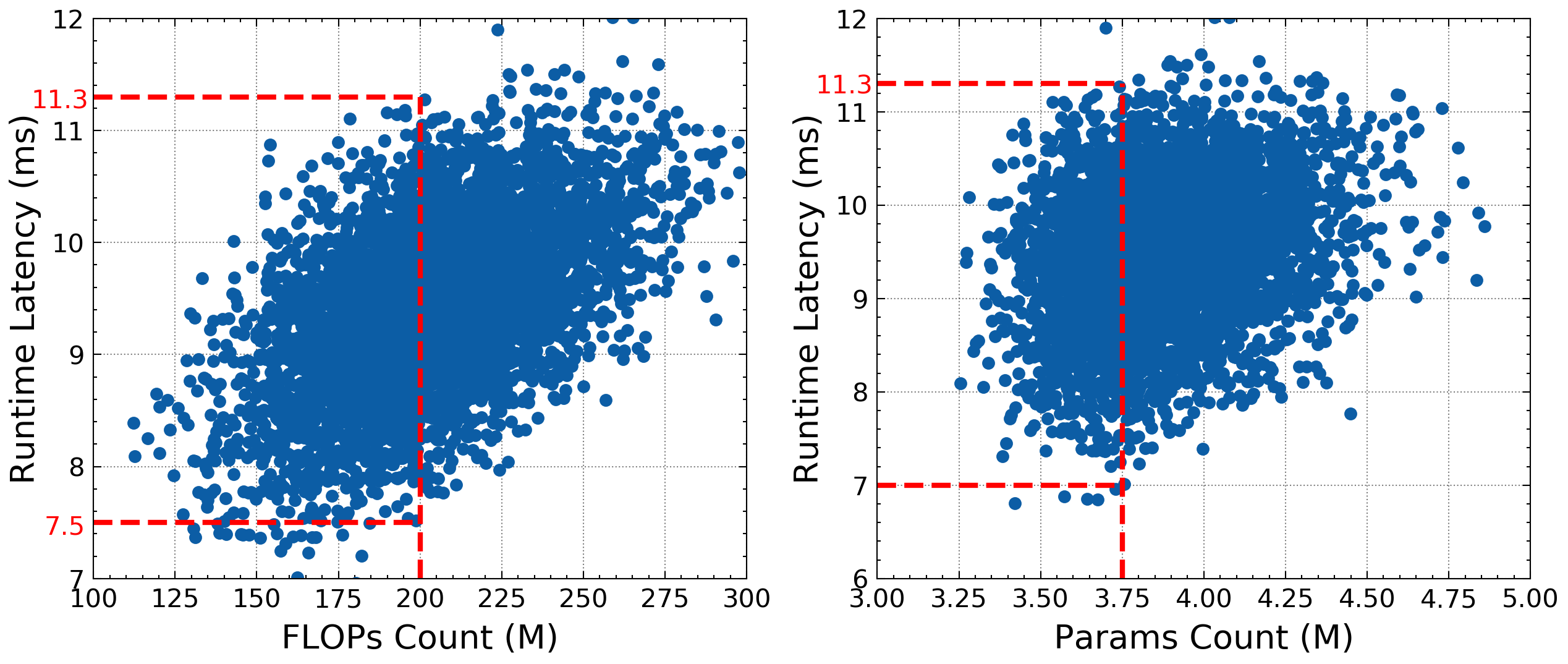}
        \vspace{-0.35cm}
    \end{center}
    \caption{Illustration of relationships between the runtime latency and FLOPs (\textit{left}) / Params (\textit{right}) count.}
\vspace{-0.60cm}
\label{fig:motivation}
\end{figure}

\section{HSCoNAS Framework}
\label{sec:methodology}
The demonstrations of the proposed HSCoNAS are twofold. From \textit{hardware's perspective}, we present an effective hardware performance modeling method to approximate the runtime latency of DNNs upon target hardware. From \textit{software's perspective}, we introduce a multi-objective evolutionary algorithm (EA) based NAS approach, where a novel dynamic channel scaling scheme is integrated to enable HSCoNAS to perform channel-level explorations. Besides, we introduce a novel progressive space shrinking method to improve the quality of the search space towards target hardware.

\subsection{Hardware Performance Modeling}
\label{sec:hardware-performance-modeling}

As illustrated in Fig. \ref{fig:motivation}, we observe that neural architectures with the same FLOPs or Params count significantly differ regarding the runtime latency. Therefore, the FLOPs or Params count is a hardware-agnostic metric and is inadequate to reflect the runtime performance upon target hardware. However, directly measuring the runtime performance on target hardware for $arch \in \mathcal{A}$ is prohibitively expensive since the search space of NAS is immensely large, \textit{e.g.}, $|\mathcal{A}| \approx 9.5 \times 10^{33}$ in HSCoNAS. To tackle this, we analytically model the runtime latency for $arch \in \mathcal{A}$ using the following formulation:
\begin{equation}
    {\scriptsize LAT(arch) = \sum_{l=1}^L op^l + \mathcal{B}}
    \label{eq:latency}
\end{equation}
where $op^l$ represents the operator of $l$-th layer in $arch$, \textit{i.e.}, $arch = \{op^l\}_{l=1}^L$. Here $\mathcal{B}$ is incorporated to compensate the communication overheads in sequential layers and can be empirically approximated as follows:
\begin{equation}
    {\scriptsize \mathcal{B} = \frac{1}{M} (\sum_{i=1}^M LAT^+(arch_i) - \sum_{i=1}^M LAT(arch_i))}
    \label{eq:latency-bias}
\end{equation}
where $LAT^+(arch_i)$ denotes the on-device runtime latency of $arch_i$. $M$ is the number of architectures sampled from $\mathcal{A}$. 

\begin{figure}[t]
    \begin{center}
        \includegraphics[width=0.80\columnwidth]{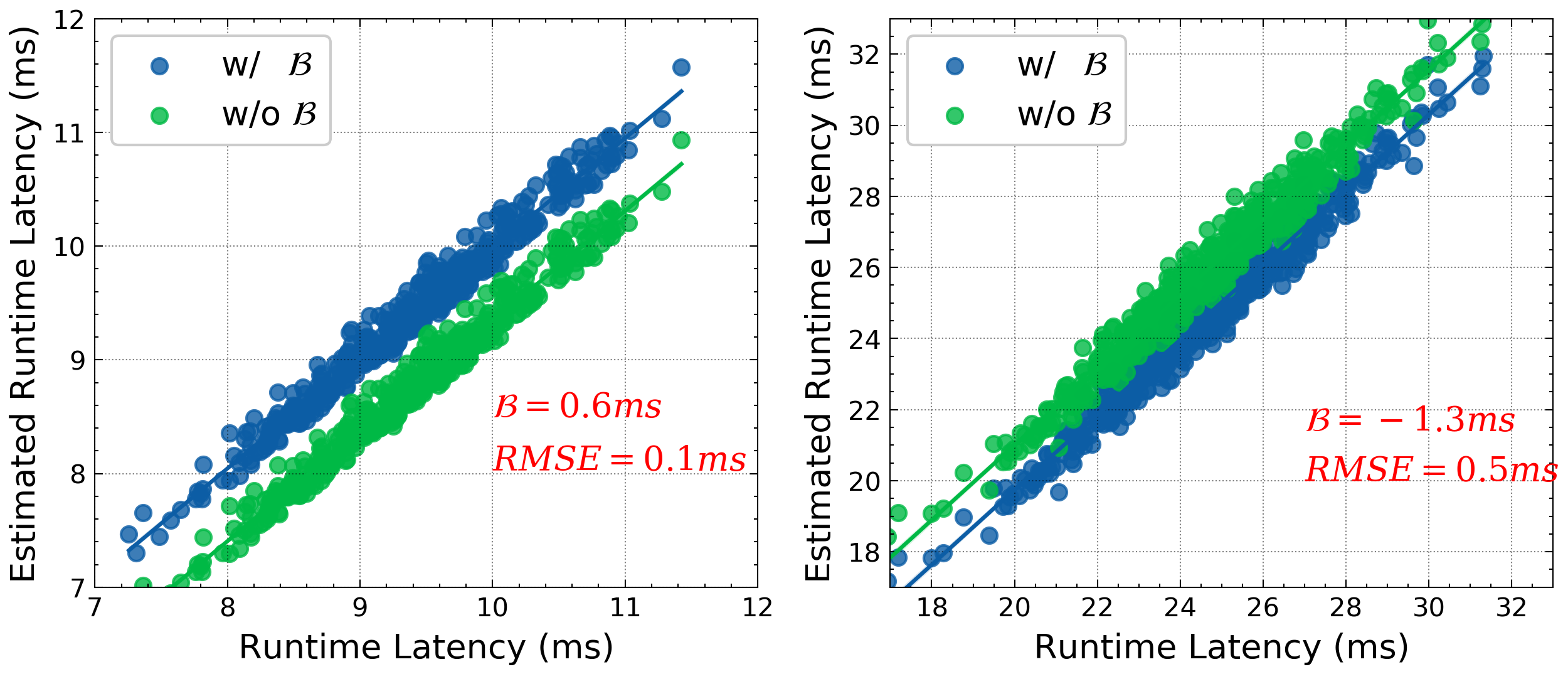}
        \vspace{-0.35cm}
    \end{center}
    \caption{Illustration of the effectiveness of proposed hardware performance modeling method on GPU (\textit{left}) and CPU (\textit{right}).}
\vspace{-0.5cm}
\label{fig:prediction-cpu-gpu}
\end{figure}

We perform experiments on three hardware devices, \textit{i.e.}, GPU (Nvidia Quadro GV100), CPU (Intel Xeon Gold 6136), and edge device (Nvidia Jetson Xavier). Please note we set the batch size as 1, 16, 32 for CPU, edge device, and GPU since small batch size will lead to resource under-utilization \cite{cai2018proxylessnas}. For the edge device, the power mode 6 is applied across this work. As illustrated in Fig. \ref{fig:prediction-cpu-gpu}, we observe a strong correlation between the on-device and the estimated runtime latency after incorporating $\mathcal{B}$. Notably, the proposed method achieves an extremely low root-mean-squared-error (RMSE) of 0.1ms, 0.5ms, 1.7ms for CPU, GPU, edge device, respectively. 

\subsection{Dynamic Channel Scaling}
\label{sec:dynamic-channel-scaling}

\begin{figure}[t]
    \begin{center}
        \includegraphics[width=0.78\columnwidth]{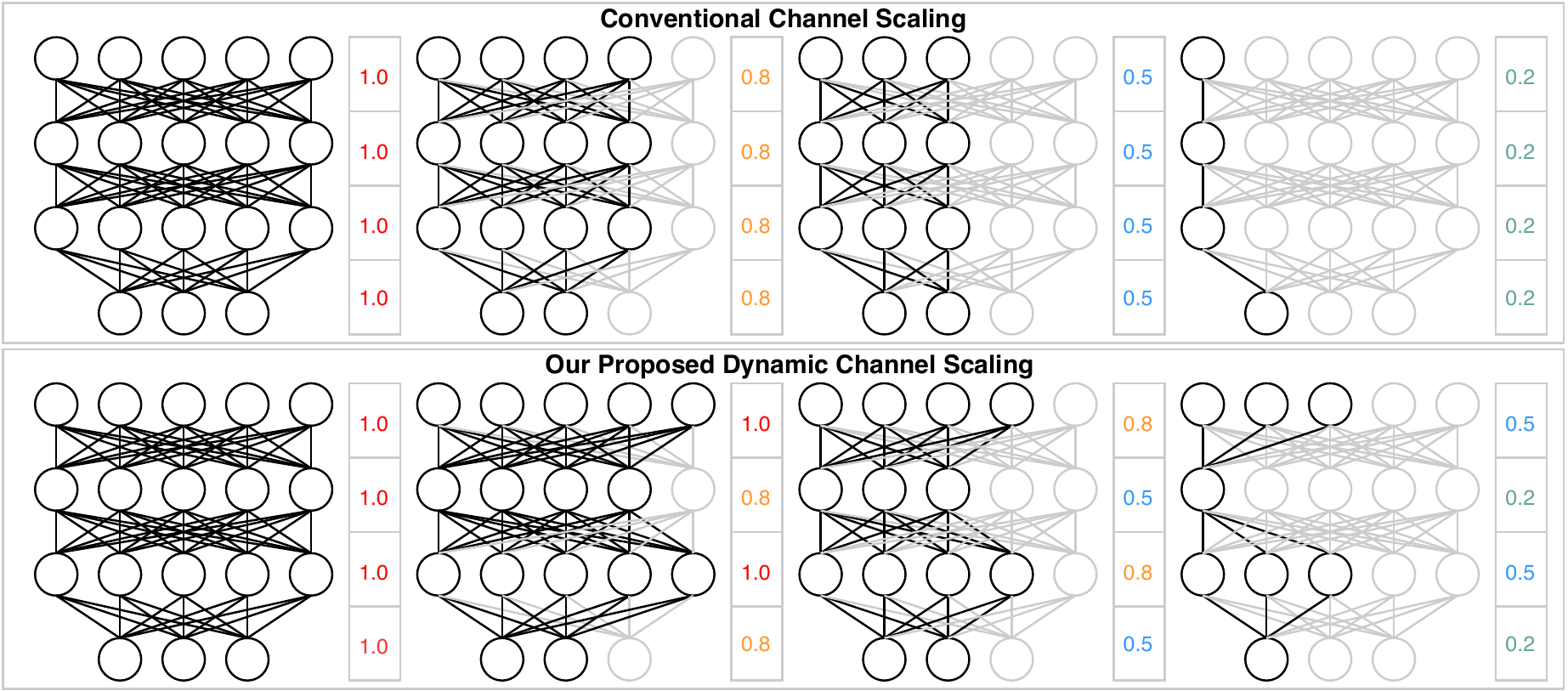}
        \vspace{-0.32cm}
    \end{center}
    \caption{Comparisons between the conventional (\textit{top}) and the proposed dynamic channel scaling (\textit{bottom}) scheme. Note the scaled number of channels is rounded (e.g., $5 \times 0.5 \approx 3$).}
    \vspace{-0.6cm}
\label{fig:channel-scaling}
\end{figure}

In literature, the hardware-aware NAS works \cite{cai2018proxylessnas, tan2019mnasnet, wu2019fbnet} merely search for the optimal configuration of the operator in each layer while keeping the number of channels in each operator fixed. However, as demonstrated in \cite{yu2018slimmable, sandler2018mobilenetv2}, the number of channels has an essential impact on both accuracy and runtime efficiency upon target hardware. Nonetheless, the conventional channel scaling scheme \cite{yu2018slimmable} is performed after the neural architecture is determined, and a uniform scaling factor is imposed across layers as illustrated in Fig. \ref{fig:channel-scaling} (\textit{top}), thereby cannot achieve effective trade-offs between accuracy and efficiency. To alleviate these issues, we propose a dynamic channel scaling scheme as depicted in Fig. \ref{fig:channel-scaling} (\textit{bottom}), which is further integrated into HSCoNAS to enable the channel-level explorations, \textit{i.e.}, determining the best channel configuration for each layer. To achieve this, we first define a list of $n$ channel scaling factors in HSCoNAS as $C = \{c_1, c_2, ..., c_n\}$, \textit{e.g.}, $\{0.1, 0.2, ..., 1.0\}$. We denote $S^l$ as the maximum number of channels for the $l$-th layer. 

Recall that NAS algorithms \cite{tan2019mnasnet, cai2018proxylessnas} select one proper operator $op^l$ from the operator set $\mathcal{O}$ for each layer $l$ to generate an architecture candidate. To begin with, we initialize the number of channels as $S^l$ for layer $l$ in the supernet, \textit{i.e.}, initialized with the maximum number. In practice, the dynamic channel scaling is implemented via scaling down from the maximum number of channels. The reason behind this is the scaling down method can avoid collapses during training the supernet since we need to reconstruct the supernet topology and reload the inherited weights into memory once the scaled number of channels is larger than the initialized one. 
Throughout training the supernet, we leverage a masking mechanism with the vector $\mathbb{I}^l \in \{0,1\}^{S^l}$, where the scaling factor $c^l \in C$ is used to manipulate the number of channels for each operator within layer $l$, \textit{i.e.}, assigning 1 to the selected channels and 0 to the masked ones. By means of the scaling down method, we can derive the output of $op^l$ as $\mathbb{I}^l \times op^l(x)$, where $x$ denotes the output of the previous layer. To incorporate the dynamic channel scaling into HSCoNAS, we adjust the supernet training accordingly, \textit{i.e.}, we change $arch \in \mathcal{A}$ from $\{op^l\}_{l=1}^L$ to $\{op^l, c^l\}_{l=1}^L$, where $c^l \in C$ is dynamically imposed across different layers. After the supernet is well trained, the proposed EA-based architecture search (refer to Section \ref{sec:architecture-search}) can automatically discover the optimal architecture candidate upon target hardware, \textit{i.e.}, $arch^*=\{op^{l*}, c^{l*}\}_{i=1}^L$, using the weight-sharing technique.

\subsection{Progressive Space Shrinking}
\label{sec:progressive-space-shrinking}

Since the search space of NAS is combinatorially large, we propose an efficient space shrinking method to progressively prune and shrink the initial search space, which finally arrives at a well-designed subspace where it is much easier to find superior DNNs upon target hardware \cite{radosavovic2019network}. Thus, we only need to explore the well-designed subspace instead of the whole space, thereby significantly improving the search efficiency. In HSCoNAS, we characterize the quality of different subspaces upon target hardware with statistical distribution estimates. 
\begin{definition}
Given a set of $N$ architecture candidates uniformly sampled from a subspace $\mathcal{A}^{sub}$, the quality $Q(\mathcal{A}^{sub})$ of $\mathcal{A}^{sub}$ upon target hardware is defined as follows:
\begin{equation}
    {\small Q(\mathcal{A}^{sub}) = \frac{1}{N} \sum_{i=1}^N \mathcal{F}(arch_i, T), \,\, \mathrm{s.t.} \,\, arch_i \sim \mathcal{U}(\mathcal{A}^{sub})}
    \label{eq:efficiency}
\end{equation}
where $\mathcal{F}(\cdot)$ represents the objective defined in Eq (\ref{eq:relaxed-objective}).
\label{def:quality}
\end{definition}
For instance, for two subspaces $\mathcal{A}^1$ and $\mathcal{A}^2$, if $Q(\mathcal{A}^1) > Q(\mathcal{A}^2)$, this indicates that we have a higher probability to find better DNNs in terms of the trade-off between latency and accuracy within subspace $\mathcal{A}^1$. We set $N$ to 100 across our experiments, which is proven to be sufficient in \cite{radosavovic2019network}.

\begin{figure}[t]
    \begin{center}
        \includegraphics[width=0.78\columnwidth]{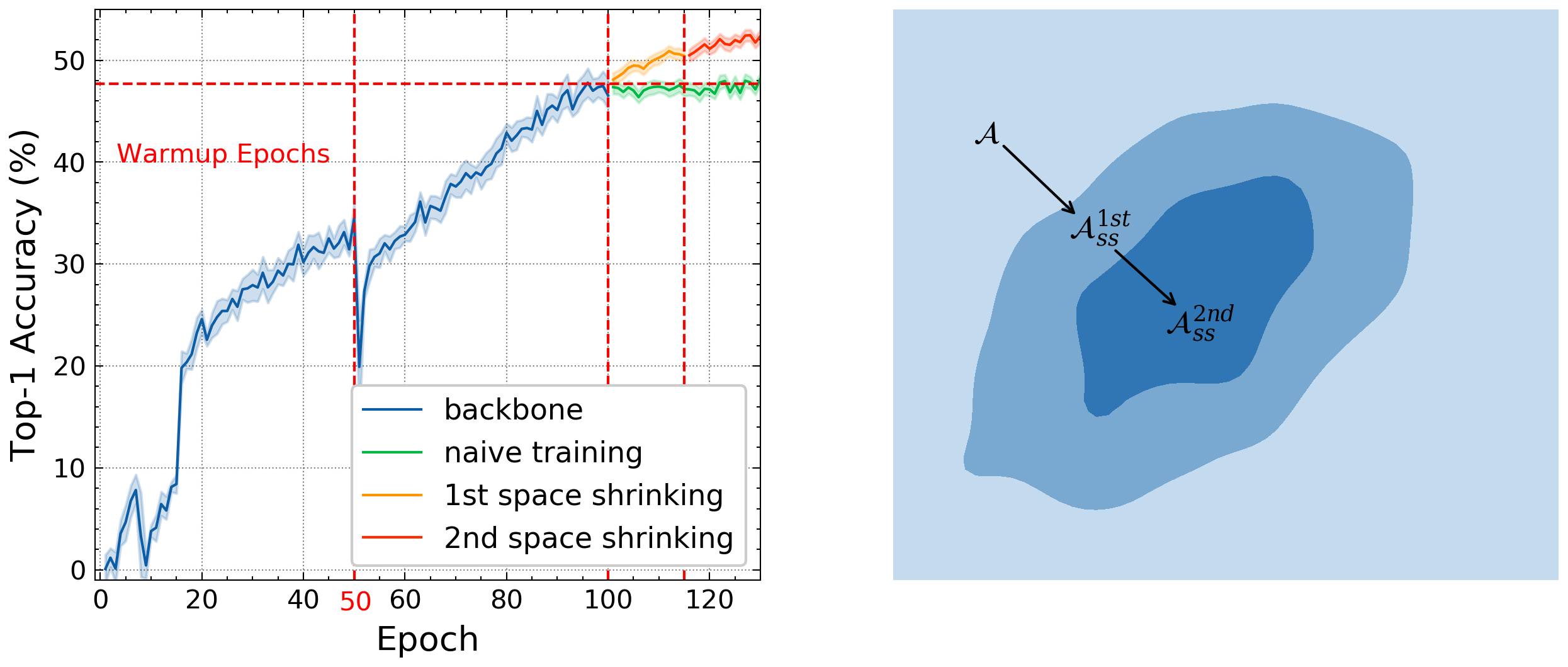}
        \vspace{-0.35cm}
    \end{center}
    \caption{Illustration of the progressive space shrinking scheme.}
    \vspace{-0.5cm}
    \label{fig:supernet-stage1-3}
\end{figure}

The progressive space shrinking consists of three stages: the initial search space $\mathcal{A}$, the first space shrinking to $\mathcal{A}_{ss}^{1st}$, and the second space shrinking to $\mathcal{A}_{ss}^{2nd}$. These procedures are illustrated in Fig. \ref{fig:supernet-stage1-3} (\textit{right}). 
To begin with, we train the supernet $\mathcal{N}$ for 100 epochs within the initial search space $\mathcal{A}$, which serves as the foundation for the subsequent space shrinking steps since we need to have architecture samples to approximate the quality of different subspaces. Then, we start the first space shrinking, where we sample a subspace for each operator within each layer and use Definition \ref{def:quality} to evaluate the quality of different subspaces. Finally, the operator with the highest quality is selected for that layer. For the first stage we apply this space shrinking to $20$-th, $19$-th, $18$-th, $17$-th, layer by layer. Note that when we evaluate the subspaces of the current layer, the operator of its subsequent layer should be fixed. For example, when evaluating the 19-th layer, we fix the operator of 20-th layer according to the subspace quality. After applying the space shrinking to the four layers, we complete the first stage space shrinking, reaching a smaller design space $\mathcal{A}_{ss}^{1st}$, which reduces the space size by three orders of magnitudes. 
Furthermore, we tune the supernet within subspace $\mathcal{A}_{ss}^{1st}$ for 15 epochs and then conduct the second space shrinking in order of $16$-th, $15$-th, $14$-th, $13$-th layer in the same way. At the end, we arrive at $\mathcal{A}_{ss}^{2nd}$, further reducing the space size by another three orders of magnitudes. 

In terms of complexity, if we evaluate the subspaces of four layers at the same time, it needs to evaluate $5^4$ subspaces, whereas our method only needs to evaluate $5\times 4$ subspaces. Moreover, as illustrated in Fig. \ref{fig:search-xavier} (\textit{left}), we observe after each space shrinking the supernet obtains higher accuracy when compared with \textit{naive training}, which indicates to continue training the supernet within the initial space $\mathcal{A}$. 

\subsection{Evolutionary Architecture Search}
\label{sec:architecture-search}

\begin{figure}[t]
    \begin{center}
        \includegraphics[width=0.78\columnwidth]{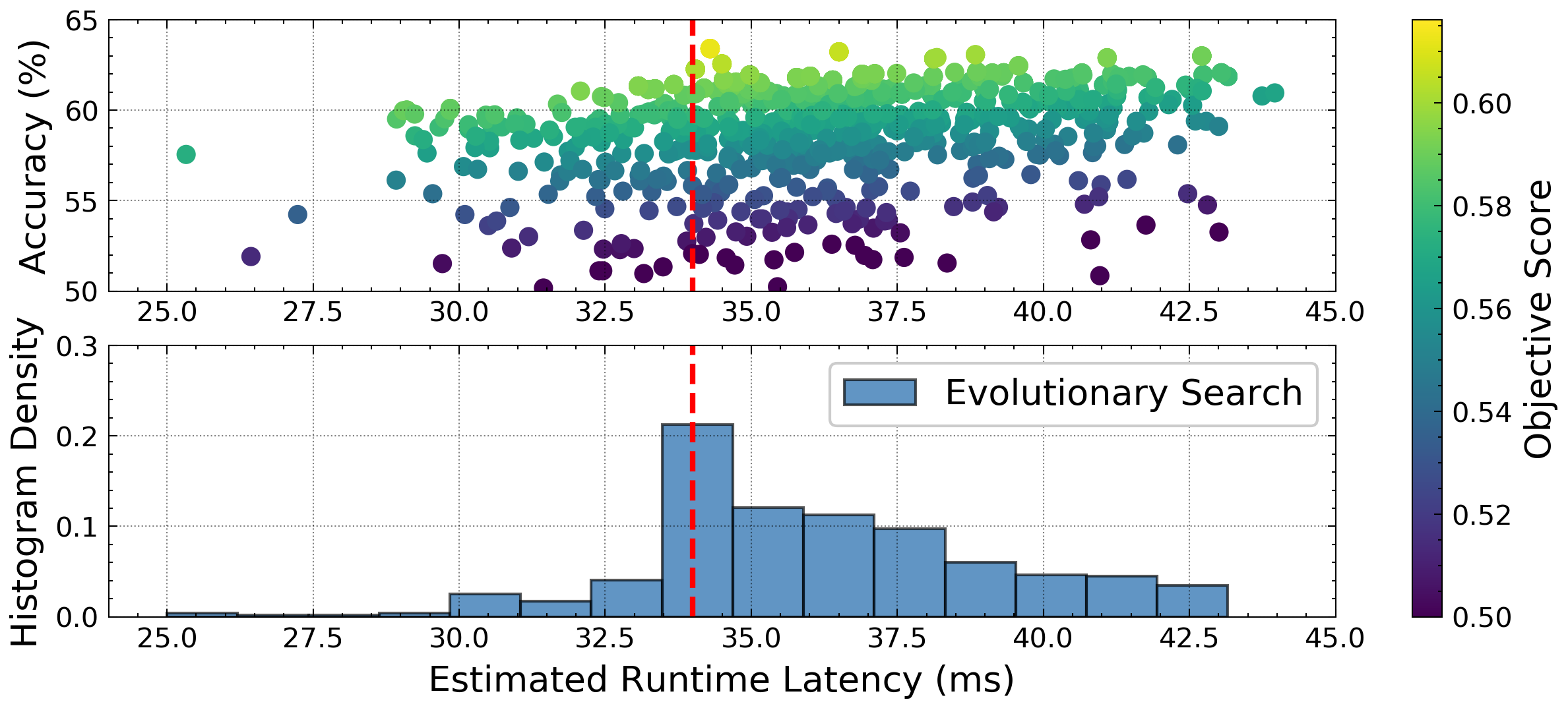}
        \vspace{-0.35cm}
    \end{center}
    \caption{Illustration of the evolutionary results. The red dashed line denotes the latency constraint for edge device, \textit{i.e.}, 34ms.}
\vspace{-0.6cm}
\label{fig:search-xavier}
\end{figure}

With all the key components introduced before, this section presents our architecture search method. EA and RL are two widely used algorithms in literature. However, RL incurs a high search cost since it is hard to converge \cite{zoph2018learning-rl}. Thus, we adopt EA across this work, which is as effective as RL but with higher efficiency \cite{real2019regularized}. HSCoNAS aims to search for the architecture candidate with the highest objective score (refer to Eq (\ref{eq:relaxed-objective})), which can be formulated as follows:
\begin{equation}
    {\small {arch}^* = \mathop{\mathrm{arg \, max}} \limits_{arch \in \mathcal{A}} \mathcal{F}(arch, T)}
    \label{eq:search}
\end{equation}
where $arch = \{op^l, c^l\}_{l=1}^L$ represents the architecture candidate sampled from the supernet with inherited weights. We set the number of generations as 20, the size of population as 50, the number of parents as 20, respectively. During each evolution, crossover with a probability of $0.25$ and mutation with a probability of $0.25$ jointly work to yield efficient explorations not only on the operator level but also on the channel level.  We take the evolutionary results on the edge device as an example, which is illustrated in Fig. \ref{fig:search-xavier} (\textit{top}), where a specified latency requirement of 34ms is given. Notably, HSCoNAS discovers an optimal neural architecture with the runtime latency of 34.3ms, which approximately meets the specified latency constraint. We visualize the results in a histogram as seen in Fig. \ref{fig:search-xavier} (\textit{bottom}), where we find EA can find more architecture candidates which have the runtime latency closed to the specified latency constraint, \textit{i.e.}, 34ms. 

\section{Experimental Settings and Results}
\label{sec:experiments}

In this section, we evaluate HSCoNAS on three hardware platforms, \textit{i.e.}, GPU (Nvidia Quadro GV100), CPU (Intel Xeon Gold 6136), edge device (Nvidia Jetson Xavier), with the specified latency constraint of 9ms, 24ms, 34ms, respectively.


\subsection{Experimental Settings}
\label{sec:experimental-settings}

In this work, we apply ImageNet as the experimental dataset. In practice, we train the supernet backbone using the SGD optimizer with a momentum of 0.9, a weight decay of $3 \times 10^{-5}$,  a norm gradient clipping of 5, a batch size of 512, a learning rate of 0.5 annealed down to zero following the cosine schedule for 100 epochs. The standard data augmentations are applied. After each stage of progressive space shrinking, the supernet is tuned within the shrunk search space for 15 epochs with an initial learning rate of 0.01 and 0.0035, respectively. We denote those architectures discovered by HSCoNAS as HSCoNets, which will be trained from scratch for fair comparisons. The training settings are the same as training the supernet backbone with two exceptions. The batch size is set to 1024 and the learning rate warm-up strategy is applied for the first five epochs. Afterward, we set the batch size as 1, 16, 32 for CPU, edge device, GPU to evaluate the runtime latency.

\subsection{Experimental Results}
\label{sec:experimental-results}
Similar to \cite{sandler2018mobilenetv2, howard2019searching}, we apply two channels layouts, \textit{i.e.}, $[48, 128, 256, 512]$ and $[68, 168, 336, 672]$, to generate two different sizes of HSCoNets denoted as HSCoNet-A and HSCoNet-B, respectively. The search space consists of building blocks of ShuffleNetV2 \cite{ma2018shufflenet} with different kernel sizes (\textit{e.g.}, $3\times 3$). Besides, a skip-connection operation is incorporated to allow flexible architecture search as seen in \cite{liu2018darts, cai2018proxylessnas, tan2019mnasnet}.

\begin{table}[t]
\caption{Comparisons with state-of-the-art approaches.}
\vspace{-0.3cm}
\label{tab:comparisons}
\begin{center}
\resizebox{0.8\columnwidth}{!}{%
\begin{tabular}{cccccc}
\hline
\hline
\multicolumn{1}{c|}{} & \multicolumn{2}{c|}{Test Error (\%)} & \multicolumn{3}{c}{Runtime Latency (ms)} \\ \hline
\multicolumn{1}{c|}{} & \multicolumn{1}{c|}{Top-1} & \multicolumn{1}{c|}{Top-5} & \multicolumn{1}{c|}{GPU} & \multicolumn{1}{c|}{CPU} & Edge \\ \hline
\multicolumn{1}{l|}{\textbf{Manually-Designed Models}} & \multicolumn{1}{c|}{} & \multicolumn{1}{c|}{} & \multicolumn{1}{c|}{} & \multicolumn{1}{c|}{} &  \\ \hline
\multicolumn{1}{c|}{MobileNetV2 $1.0\times$ \cite{sandler2018mobilenetv2}} & \multicolumn{1}{c|}{28.0} & \multicolumn{1}{c|}{-} & \multicolumn{1}{c|}{11.5} & \multicolumn{1}{c|}{25.2} & 61.9 \\ \hline
\multicolumn{1}{c|}{ShufflenetV2 $1.5\times$ \cite{ma2018shufflenet}} & \multicolumn{1}{c|}{27.4} & \multicolumn{1}{c|}{-} & \multicolumn{1}{c|}{10.5} & \multicolumn{1}{c|}{34.3} & 65.9 \\ \hline
\multicolumn{1}{c|}{MobileNetV3 (\textit{large}) \cite{howard2019searching}} & \multicolumn{1}{c|}{24.8} & \multicolumn{1}{c|}{-} & \multicolumn{1}{c|}{12.2} & \multicolumn{1}{c|}{31.8} & 61.1 \\ \hline
\multicolumn{6}{l}{\textbf{State-of-the-art NAS Models}} \\ \hline
\multicolumn{1}{c|}{DARTS \cite{liu2018darts}} & \multicolumn{1}{c|}{26.7} & \multicolumn{1}{c|}{8.7} & \multicolumn{1}{c|}{17.3} & \multicolumn{1}{c|}{81.4} & 68.7 \\ \hline
\multicolumn{1}{c|}{MnasNet-A1 \cite{tan2019mnasnet}} & \multicolumn{1}{c|}{24.8} & \multicolumn{1}{c|}{7.5} & \multicolumn{1}{c|}{10.9} & \multicolumn{1}{c|}{26.4} & 51.8 \\ \hline
\multicolumn{1}{c|}{FBNet-A \cite{wu2019fbnet}} & \multicolumn{1}{c|}{27.0} & \multicolumn{1}{c|}{9.1} & \multicolumn{1}{c|}{10.5} & \multicolumn{1}{c|}{21.6} & 48.6 \\ \hline
\multicolumn{1}{c|}{FBNet-B \cite{wu2019fbnet}} & \multicolumn{1}{c|}{25.9} & \multicolumn{1}{c|}{8.2} & \multicolumn{1}{c|}{13.6} & \multicolumn{1}{c|}{25.5} & 57.1 \\ \hline
\multicolumn{1}{c|}{FBNet-C \cite{wu2019fbnet}} & \multicolumn{1}{c|}{25.1} & \multicolumn{1}{c|}{7.7} & \multicolumn{1}{c|}{15.5} & \multicolumn{1}{c|}{28.7} & 66.4 \\ \hline
\multicolumn{1}{c|}{ProxylessNAS-GPU \cite{cai2018proxylessnas}} & \multicolumn{1}{c|}{24.9} & \multicolumn{1}{c|}{7.5} & \multicolumn{1}{c|}{12.0} & \multicolumn{1}{c|}{24.5} & 57.4 \\ \hline
\multicolumn{1}{c|}{ProxylessNAS-CPU \cite{cai2018proxylessnas}} & \multicolumn{1}{c|}{24.7} & \multicolumn{1}{c|}{-} & \multicolumn{1}{c|}{16.1} & \multicolumn{1}{c|}{29.6} & 70.1 \\ \hline
\multicolumn{1}{c|}{ProxylessNAS-Mobile \cite{cai2018proxylessnas}} & \multicolumn{1}{c|}{25.4} & \multicolumn{1}{c|}{7.8} & \multicolumn{1}{c|}{11.5} & \multicolumn{1}{c|}{26.4} & 53.5 \\ \hline
\multicolumn{6}{l}{\textbf{Hardware-Aware Models Discovered by HSCoNAS}} \\ \hline
\multicolumn{1}{c|}{HSCoNet-GPU-A} & \multicolumn{1}{c|}{\textbf{25.1}} & \multicolumn{1}{c|}{7.7} & \multicolumn{1}{c|}{\textbf{9.0}} & \multicolumn{1}{c|}{26.5} & 43.4 \\ \hline
\multicolumn{1}{c|}{HSCoNet-CPU-A} & \multicolumn{1}{c|}{25.3} & \multicolumn{1}{c|}{\textbf{7.6}} & \multicolumn{1}{c|}{10.1} & \multicolumn{1}{c|}{\textbf{22.8}} & 43.1 \\ \hline
\multicolumn{1}{c|}{HSCoNet-Edge-A} & \multicolumn{1}{c|}{25.7} & \multicolumn{1}{c|}{8.1} & \multicolumn{1}{c|}{9.9} & \multicolumn{1}{c|}{25.8} & \textbf{34.9} \\ \hline
\multicolumn{1}{c|}{HSCoNet-GPU-B} & \multicolumn{1}{c|}{23.6} & \multicolumn{1}{c|}{6.9} & \multicolumn{1}{c|}{\textbf{12.0}} & \multicolumn{1}{c|}{31.6} & 76.9 \\ \hline
\multicolumn{1}{c|}{HSCoNet-CPU-B} & \multicolumn{1}{c|}{\textbf{23.5}} & \multicolumn{1}{c|}{\textbf{6.8}} & \multicolumn{1}{c|}{13.4} & \multicolumn{1}{c|}{\textbf{26.4}} & 69.1 \\ \hline
\multicolumn{1}{c|}{HSCoNet-Edge-B} & \multicolumn{1}{c|}{23.8} & \multicolumn{1}{c|}{6.9} & \multicolumn{1}{c|}{12.9} & \multicolumn{1}{c|}{31.8} & \textbf{52.7} \\ \hline
\hline
\end{tabular}%
}
\end{center}
\vspace{-0.7cm}
\end{table}

Results and comparisons with recent state-of-the-art works are summarized in TABLE \ref{tab:comparisons}. Please note HSCoNet-Edge-A denotes the hardware-aware DNNs for the edge device, and vice versa. Remarkably, the HSCoNets outperform those manually-designed lightweight DNNs like ShuffleNetV2 \cite{ma2018shufflenet} and MobileNetV2 \cite{sandler2018mobilenetv2} in terms of both accuracy and runtime latency on our three experimental hardware devices, respectively. Besides, HSCoNet-GPU-A obtains comparable accuracy as ProxylessNAS-GPU on ImageNet while being x1.3 faster on GPU. Besides, HSCoNet-GPU-B maintains similar runtime latency on GPU as ProxylessNAS-GPU but achieves +1.3\% higher accuracy. Furthermore, HSCoNet-CPU-B yields the lowest top-1/5 error of 23.5\%/6.8\% among those state-of-the-art DNNs but with an inference speedup of x3.1 on CPU when compared with the hardware-agnostic NAS, \textit{i.e.}, DARTS \cite{liu2018darts}, which demonstrates the effectiveness of our hardware-software co-design paradigm. 

\section{Conclusion}
\label{sec:conclusion}

In this work, we introduce a hardware-aware evolutionary algorithm (EA) based neural architecture search (NAS) framework, dubbed HSCoNAS. HSCoNAS is integrated with an effective hardware performance modeling method and two NAS improvements to automate the design of efficient deep neural networks (DNNs) upon target hardware devices. Extensive experimental results demonstrate HSCoNAS outperforms recent state-of-the-art methods in terms of latency and accuracy. In future, we plan to extend HSCoNAS, which will incorporate different hardware constraints like power consumption.

\bibliography{manuscript}

\end{document}